\documentclass{article}

    \usepackage[nonatbib,preprint]{neurips_2022}

\usepackage[utf8]{inputenc} 
\usepackage[T1]{fontenc}    
\usepackage{url}            
\usepackage{booktabs}       
\usepackage{amsfonts}       
\usepackage{nicefrac}       
\usepackage{microtype}      
\usepackage{xcolor}         

\usepackage{graphicx}
\usepackage{subcaption} 
\usepackage{amsmath,amssymb,amstext}
\usepackage{algorithm}
\usepackage{algpseudocode}
\usepackage{array}              
\newcolumntype{L}{>{$}l<{$}}    
\newcolumntype{C}{>{$}c<{$}}    
\newcolumntype{R}{>{$}r<{$}}    

\DeclareMathOperator*{\argmin}{arg\,min}
\DeclareMathOperator*{\argmax}{arg\,max}

\title{Focused Adversarial Attacks}

\author{%
    Thomas Cilloni, Charles Walter, Charles Fleming\\
    Department of Computer and Information Science \\
    University of Mississippi, USA \\
    \texttt{tcilloni@go.olemiss.edu, cwwalter@olemiss.edu , fleming@olemiss.edu}
}

\begin{document}
\maketitle

\begin{abstract}
Recent advances in machine learning show that neural models are vulnerable to minimally perturbed inputs, or adversarial examples. Adversarial algorithms are optimization problems that minimize the accuracy of ML models by perturbing inputs, often using a model's loss function to craft such perturbations. State-of-the-art object detection models are characterized by very large output manifolds due to the number of possible locations and sizes of objects in an image. This leads to their outputs being sparse and optimization problems that use them incur a lot of unnecessary computation.

We propose to use a very limited subset of a model's learned manifold to compute adversarial examples. Our \textit{Focused Adversarial Attacks} (FA) algorithm identifies a small subset of sensitive regions to perform gradient-based adversarial attacks. FA is significantly faster than other gradient-based attacks when a model's manifold is sparsely activated. Also, its perturbations are more efficient than other methods under the same perturbation constraints. We evaluate FA on the COCO 2017 and Pascal VOC 2007 detection datasets. Code can be found at <see supplementary material>.
\end{abstract}

\section{Introduction}

Recent research in machine learning lead to the discovery of \textit{adversarial examples}, which are data points carefully tailored to induce errors in ML models. Many adversarial attacks likely take place every day without being detected or leaving traces. The introduction of Generative Adversarial Networks (GANs) \cite{gan} gave rise to Deepfakes, synthetic videos, pictures, and voice recordings, which pose big threats such as fake news and crafted material. Detectors have been developed to distinguish real from synthetic images \cite{deepfake_detectors}, but these have also been shown to be vulnerable to adversarial examples \cite{deepfake_detectors_fool}. Adversarial examples can also be used to promote user privacy, as shown in \cite{fawkes} and \cite{ulixes} for unauthorized facial recognition, and \cite{prevent_inference} to prevent data inference from a model's parameters.

Most adversarial examples focus on fooling Object Detection models. Object Detection is the task of localizing and classifying objects in images. Most object detection models nowadays use deep CNNs and learn feature mappings that can be post-processed to produce detections. \textit{Region Proposal Networks} (RPNs) first find regions in the image where objects may reside, and then try to predict a label for each region, possibly rejecting any proposals. This approach allows the model to be flexible in predicting large, small, or numerous objects but requires significant computations. Examples of this architecture are RetinaNet \cite{retinanet} and Faster R-CNN \cite{frcnn}.
\textit{Single Shot Detectors} (SSDs), on the other hand, layout sparse and dense grids of anchor points over images. Then, for each point in each grid, a number of probability distributions over classes are computed. This feature map is processed with \textit{non-max suppression} to discard overlapping and non-confident predictions. While faster, this method is less accurate than RPNs. An example of this architecture is SSD \cite{ssd300}. SSDs and RPNs both output very large, sparsely activated feature maps, that induce unnecessary computation in finding adversarial examples. By using only a subset of the outputs, we show how adversarial examples can be more effectively generated.

Fooling an Object Detection model with an adversarial example is defined as an adversarial attack. Adversarial attacks are algorithms that generate adversarial examples, crafted data points that are classified by neural networks differently than how they would be classified by a human. Adversarial attacks, given a real data point and an ML model, try to solve two problems at the same time: maximize the error of the model, and minimize the change introduced in the data point. The taxonomy of adversarial attacks identifies as \textit{targeted} attacks those whose goal it to fool a model into predicting a certain target; otherwise, an attack is \textit{untargeted}. \textit{White-} and \textit{black-box} methods refer to whether in the threat model attackers have access to the victim network or not.

Understanding adversarial attacks is the first step toward building defense mechanisms, and in this paper, we provide new insights into their generation. We propose a novel strategy to generate adversarial examples for object detectors that takes advantage of the semantics of the feature maps that model learns. Tracking only highly activated outputs through a neural network back to the input images, it is possible to generate adversarial cloaks that target only sensitive regions of images. These cloaks are not only less perceptible, but also more effective and equally fast or faster than comparable methods. We evaluate our strategy using three state-of-the-art object detectors: RetinaNet, SSD300, and Faster R-CNN and two datasets: COCO 2017 and VOC2007. Our results show that our algorithm is computationally more efficient and has higher efficacy for a given level of perturbation than other gradient or Generative Adversarial Network (GAN) based attacks.

\section{Related Work}
\label{sec:literature}

One of the early works in adversarial attacks against object detectors is the Dense Adversary Generator (DAG) by Xie et al. \cite{dag}. It proposes to apply gradient-based attacks previously used against classification models in the context of object detectors and semantic segmentation. Results show that the accuracy (mean Average Precision) of object detection can be reduced by DAG from $\approx70\%$ to below $20\%$. Adversarial examples generated in white-box settings are also shown to transfer to the same detection models trained on different data, however tranferability to different models is low.

Gu et al. \cite{lit:shielding} propose \textit{Gradient Shielding}, a gradient-based attack for image classifiers that is targeted at sensitive regions of images. Such selection is made at the image level, either manually (Interactive Gradient Shielding, IGS), or automatically (Automatic Gradient Shielding, AGS). IGS can be visualized as a square-shaped adversarial perturbation applied to a square region of the image smaller than the image itself; in other words, IGS ignores image borders. AGS automates this process by finding sensitive regions beforehand and zeroing gradients for insensitive regions. Our work follows the same concept as gradient shielding, that is applying perturbations only to sensitive regions, however with some key differences: the loss function used in training the victim model is irrelevant to us; the method to find sensitive regions uses learned feature maps at the output level, instead of the loss function's gradient at the input level, which further optimizes the gradient calculation and leads to more strongly activated gradients; we apply our method to object detectors, which have much larger feature maps and complex gradients, instead of image classifiers.


Wei et al. \cite{uea} propose an adversarial algorithm that uses GANs and a multi-scale attention feature loss to produce adversarial examples that can fool object detectors and classifiers alike: Unified and Efficient Adversary (UEA). This loss function is an ensemble of a high-level class loss and a low-level feature loss, and is used in training the GAN's generator. Their results show a drop in the accuracy of Faster R-CNN and SSD300 on the Pascal VOC 2007 dataset from 70\% to 5\% and from 68\% to 20\%, respectively.

A drawback of UEA is the time it takes to train the generator. Given a train set of a few thousand samples, the generator requires several days of training time. 
More Imperceptible attacks (MI) \cite{mi} overcome this problem by improving upon the generator used in UEA \cite{uea} with an early stopping condition and a noise reduction step. Instead of iterating the generator a fixed number of times, an object detector is used to determine when an example has become adversarial, and the generator is then stopped. This ensures that generated samples are adversarial, and also minimally perturbed. The efficacy of attacks is at least as good as that of Wei et al, but at a fraction of the time ($8.4s$ to $1.8s$).

\section{Focused Adversarial Attacks}
\label{sec:method}

This section introduces the algorithm to generate focused adversarial examples. We first present some characteristics of object detectors that motivate the design of focused adversarial attacks. Then we introduce the algorithm itself, and finally provide some considerations on its implementation.

\subsection{Inspecting Object Detectors}
Object detection neural networks process images and produce a feature output that contains spacial and semantic information about any objects in the image. Contrary to neural network classifiers, which output as many features as there are classes in the task they solve (typically with a softmax layer), object detectors' outputs are much larger.

As images may contain multiple objects, and the objects may have different scales, and be of different types, the feature map of object detectors must contain all such information. Determining the class of an object is done in the same manner a classifier makes predictions: generating a probability distribution over classes, which is a one-dimensional vector. Objects have drastically different shapes, and this is partially handled by having either multiple candidate bounding boxes or dynamic bounding boxes. Minor improvements to the fit of bounding boxes are also typically controlled with a scale and offset adjustment. The spacial information of detections is represented by a grid of candidate locations in an image, and to support small and large-sized objects, multiple grids are employed. Two of the most popular object detectors, SSD\cite{ssd300} and RetinaNet\cite{retinanet}, have around 8K and 100K candidate boxes, respectively. Each candidate box has a related probability distribution over target classes, which we can assume to be COCO's 80 classes. The total size of the output of these networks, therefore, becomes enormous: 640K in SSD and 8M in RetinaNet.

The first consideration of traditional adversarial machine learning methods is to be made with regard to the intuition behind the effect of perturbations on predictions. In classification models, the gradient of the loss function, computed with respect to the input and for a single output feature, intuitively points perturbations in the direction to optimally disrupt that feature or class. Similarly, considering a distribution over target classes, perturbations can either confuse a model by spreading the distribution or fool it by targeting a feature other than the target. In object detectors, however, it is unclear what gradient-based attacks actually do. Given the high dimensionality of the detection features, targeted attacks can incur in racing conditions where gradients with respect to input pixels cancel each other out.

\begin{figure}
    \begin{subfigure}[t]{0.35\textwidth}
        \includegraphics[width=\columnwidth]{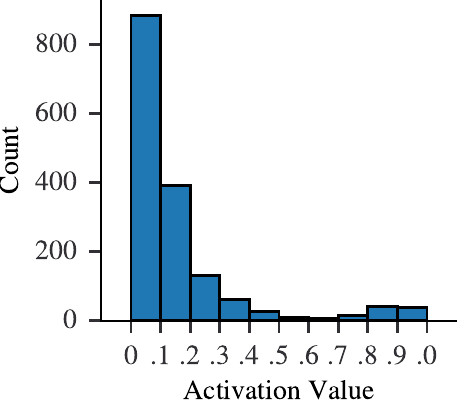}
        \caption{Histogram of the distribution of activations in the 99.98\% most activated subset of feature mappings. These are generated by feeding sample $885$ from COCO-2017-val to RetinaNet.}
        \label{fig:distribution}
    \end{subfigure}
    \hfill
    \begin{subfigure}[t]{0.6\textwidth}
        \includegraphics[width=\columnwidth]{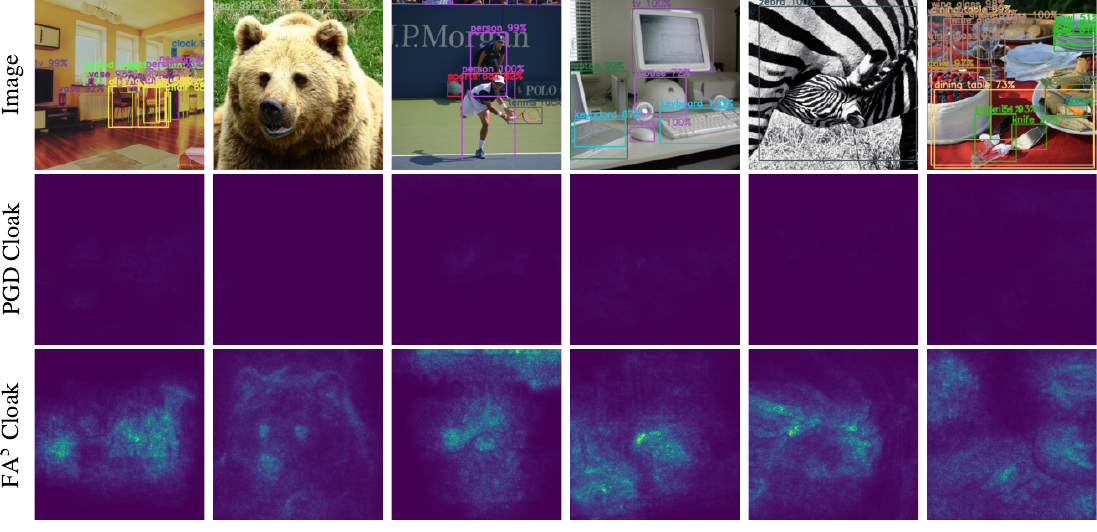}
        \caption{Heatmap of the sensitive regions of images found by the focused adversarial examples algorithm (bottom row), compared to the regions affected by standard PGD (middle row). Heatmaps are found with 5 iterations for each algorithm and all parameters are the same.}
        \label{fig:sensitive_regions}
    \end{subfigure}

    \caption{Investigative study of the activations found in Faster R-CNN' feature map.}
    \label{fig:exploration_figures}
\end{figure}

A second consideration is the characteristics of object detectors' feature mappings. The large outputs of detectors are very sparsely activated. Significant detections usually occur only in extremely small subsets of the feature maps, and they become even fewer when they are filtered with a confidence threshold. We explored this behavior by feeding an example image to RetinaNet and studying its output mapping. 99.98\% of the 8M outputs had activations at or below 0.05, showing how sparse the features are. Fig.\ref{fig:distribution} shows the distribution of the 0.02\% most activated outputs: only a very small subset of this already minuscule set would actually contribute to candidate box selection algorithms at later stages (which typically filter out activations below 0.5).

These insights are some of the motivations that led to our design of Focused adversarial attacks. We show that adversarial examples can be generated more efficiently and effectively by filtering out non-contributing parts of the feature maps of object detectors.

\subsection{Problem Statement}
Focused adversarial attacks are a form of gradient-based attacks and therefore require white-box access to the victim model. We assume that the attacker has access to the structure and the parameters of the model to attack, and can replicate the data pre-processing pipeline used in training. The loss function used in training does not need to be known. Finally, it is assumed that the model outputs the feature map of activation of classes at various locations in the image.

In object detection tasks, images $\mathbf{x} \in \mathcal{X}$ are sampled from an unknown distribution $\mathcal{X} \subset [0, 1] ^ {H \times W \times 3}$ and have corresponding one-hot encoded labels $\mathbf{y} \in \mathcal{Y} \subset [0, 1] ^ {A \times C}$ where $\mathbf{y}_ij = 1$ indicates that there is an instance of object $j$ at the anchor location $i$ (typically if $j=0$ there is no object or it is just background).
An object detection model $f$ with parameters $\theta$ maps images onto a feature space $\mathcal{Y}$ as $\mathbf{\hat y} = f(\mathbf{x}; \theta) \in \mathcal{Y}$, such that $\argmax_j \mathbf{\hat y}_{ij}$ is the predicted class for location $i$, and its confidence is $\mathbf{\hat y}_{ij}$. 
The position $i$ in a vector $\mathbf{\hat y}$ tells the location of a detection, and its coordinates in an image can be calculated taking into consideration the number of grids in the image, the size of each grid, and the number of anchors. Focused adversarial attacks, however, being model-agnostic, are not concerned with the location of detections and therefore the details of their implementation can be safely ignored.

We look at adversarial attacks as optimization problems. Images are perturbed with a minimal mask $\delta \in [0, 1] ^ {H \times W \times 3}$ with the goal of removing all confident predictions from a model's feature map. If all $\mathbf{\hat y}_{ij}$ in a model's output are low enough (the exact value depending on model implementation and choices in interpreting results), then the model will not detect any object in an image. We indicate this upper bound on ignored detections as $c$, and define the optimization problem to find perturbations $\delta$ as follows: 
\begin{equation}
\begin{aligned}
    \text{find} & \quad \argmin_\delta \| \mathbf{x} + \delta \|_\infty 
    \\
    \text{s.t.} & \quad \mathbf{\hat{y}}_{ij} \le c, \forall\; \mathbf{\hat{y}}_{ij} \in \mathbf{\hat y} = f(\mathbf{x} + \delta; \theta)
\end{aligned}
\end{equation}

\subsection{Solution Algorithm}
The novelty of our work is the method to find the perturbation $\delta$. We introduce a \textit{focus threshold} $t$ that determines whether a feature of a model's output map should be considered or not to compute the perturbations. The \textit{Focused Activation} $FA$ is defined as the L1 norm of the vector of regions of the feature map that exceed the threshold, and we minimize it with respect to input images to determine the perturbations $\delta$. $L$ and $C$ are the numbers of locations in a feature map and the classes in the detection task, and $\epsilon$ is the modulator of the intensity of perturbations.
\begin{equation}
\begin{gathered}
    FA (\mathbf{\hat y}, t) = \sum_i^A \sum_j^C \max(0, \mathbf{\hat{y}}_{ij} - t)
    \\
    \delta = \epsilon * sign \left( \nabla_\mathbf{x} FA(f(\mathbf{x}; \theta), t) \right)
\end{gathered}
\end{equation}

The motive behind filtering features is that the majority of detection feature maps are noise. By focusing only on significantly activated regions of the maps, we are able to generate perturbations that specifically target sensitive regions in an image. This drastically reduces the intensity of perturbations in background regions of images and cloaks sensitive regions more effectively. Figure \ref{fig:sensitive_regions} shows how in practice focused attacks are more targeted to sensitive image parts for object detection.

In order to make the focused activation function differential, we propose two viable solutions. The first, $FA_P$, takes advantage of the parallel processing capabilities of modern processors to speed up computations, and the second, $FA_I$, uses indexing to reduce the volume of operations to execute, which is particularly effective on sparsely activated features maps. For the remainder of this paper, where the implementation details are not specified, $FA_I$ is assumed to be the one used, because we found it to be faster than $FA_P$ in our experiments. It is advisable to always compare the performance of the two, as either one could be faster depending on the size of feature maps, the distribution of activations, and the hardware used.
\begin{equation}
\begin{gathered}
\label{eq:fap}
focus(v, t) := 
    \begin{cases}
      1 & \text{if } v > t \\
      0 & \text{otherwise}
    \end{cases} 
\\
FA_P (\mathbf{\hat y}, t) = \sum_i^A \sum_j^C \mathbf{\hat{y}}_{ij} * focus(\mathbf{\hat{y}}_{ij}, t)
\end{gathered}
\end{equation}

\begin{equation}
\begin{gathered}
\label{eq:fai}
sub (\mathbf{\hat y}, t) := \left\{ \mathbf{\hat{y}}_{ij} : \mathbf{\hat{y}}_{ij} > t, \forall\; i,j : 0 \leq i \leq L, 0 \leq j \leq C \right\} \\
FA_I (\mathbf{\hat y}, t) = \| sub (\mathbf{\hat y}, t) \|_1
\end{gathered}
\end{equation}

As with all gradient-based attacks, our focused adversarial attacks can be executed in either a one-shot or iterative fashion. Following is the iterative version of the one-shot attack proposed earlier to compute adversarial perturbations. For the remainder of this paper, this is the exact formulation that we use in all experiments, with $FA = FA_I$ and varying number of iteration steps and per-step magnitude $\epsilon$. Following is its definition, and the complete process is shown in Algorithm \ref{alg:focused_attack}.
\begin{equation}
\delta^{t+1} = \epsilon * sign \left( \nabla_\mathbf{\mathbf{x}+\delta}\; FA(f(\mathbf{x}+\delta; \theta), t) \right)
\end{equation}

\begin{algorithm}
\caption{An algorithm with caption}
\label{alg:focused_attack}

\hspace*{\algorithmicindent} \textbf{Input:} an image $\mathbf{x}$; a model $f$ with parameters $\theta$ whose outputs lie on $\mathcal{Y} \subset [0, 1]^{A \times C}$; the perturbation radius $\varepsilon \in [0,1]$; the number of iterative steps $S$; the focusing threshold $t$. 
\\
\hspace*{\algorithmicindent} \textbf{Output:} an adversarial image $\mathbf{x'}$

\begin{algorithmic}[1]
    \Function{Focused Attack}{$\mathbf{x}, f, \theta, \varepsilon, S, t$}
        \State $\epsilon \gets \varepsilon / s$         
        \Comment{Apply $\epsilon$ for $S$ steps}
        \State $\mathbf{x'} \gets \mathbf{x}$
        \For {$i \gets 1 \dots S$}
            \State $\delta \gets sign \left( \nabla_\mathbf{x'} FA_P(f(\mathbf{x'}; \theta), t) \right)$    
            \Comment{Implemented in Eq. \ref{eq:fap}}
            \State $\mathbf{x'} \gets \mathbf{x'} + \epsilon * \delta$  
            \Comment{Ensures $\|\delta\|_\infty \leq \epsilon$, thus $\| \mathbf{x'} - \mathbf{x} \|_\infty \leq \varepsilon$}
        \EndFor
        \State \Return $\mathbf{x'}$
    \EndFunction
\end{algorithmic}
\end{algorithm}

\section{Exploratory Experiments}
\label{sec:explore}

In this section, we propose an exploration of three hyperparameters of our methods, which are the total intensity of perturbations, the granularity of perturbation steps, and the focusing threshold used in the Focused Activation function. Finally, we study the performance gain of our focused adversarial attacks in terms of processing speed.

The experiments carried out in this section use a single subset of 500 samples taken uniformly at random from the COCO 2017 validation split, and Faster R-CNN as the object detection model. Efficacy is measured with the mean Average Precision (mAP) metric, and because our intent is to hinder object detection models, lower precision values indicate that a method is more effective.

\subsection{Varying $\varepsilon$}{Epsilon}-ball
The first series of experiments investigates how the intensity of perturbations affects precision. Adversarial examples are defined as $\mathbf{x'} = \mathbf{x} + \delta$, and $\delta$ is calculated by optimizing the FA function. Considering using Projected Gradient Descent, the total perturbation $\delta$ found after $T$ steps is found iteratively as the summation of $T$ $\epsilon$-magnitude, image-shaped vectors. In order to produce perturbations comparable across image samples and detection models used, we set an upper bound on the $L_\infty$ norm of the total perturbation vector $\varepsilon$, such that $\| \delta \|_\infty \leq \varepsilon$. Given $T$ iterations of the algorithm, this constraint can be guaranteed by setting $\epsilon = \varepsilon / T$. The $\varepsilon$-ball of perturbations is therefore the space in which all allowable perturbations reside.

Figure \ref{fig:explore_epsilon} shows how enlarging the $\varepsilon$-ball of perturbations makes attacks more effective. This behavior is expected as larger perturbations are likely to be more effective, though more noticeable. Results are presented for the Fast Gradient Sign Method (FGSM) and Projected Gradient Descent (PGD), used as reference scores, while our attacks are $FA^S$, with $S$ being the number of iterations of the algorithm.

\subsection{Perturbation Granularity}
The second series of experiments on the hyperparameters of focused adversarial attacks is concerned with the granularity of perturbations. Given a constant $\varepsilon$-ball of adversarial changes, a different number of iterative steps $T$ can lead to different results. The granularity of perturbations is therefore the small epsilon that determines the magnitude of each iteration's perturbation, which is inversely proportional to the number of steps as $\epsilon = \varepsilon / T$.

The precision of Faster R-CNN on adversarial examples generated with varying steps is shown in Figure \ref{fig:explore_steps}. Given a fairly large perturbation space of $0.1$, the effectiveness of attacks increases from 1 to 3 steps, and then shows drastically diminishing returns for finer-grained perturbations. On a similar note, when the perturbation space is more constrained, more than 3 steps also cause diminishing returns, and more than 5 steps even show a decrease in performance. It is, therefore, safe to assume that many but small perturbations do not perform as well as few, larger ones.

\subsection{Focusing Factor}
The last series of experiments investigates the focusing threshold used in the focused adversarial attacks. This hyper-parameter dictates which activations in a model's feature map should be considered and which excluded. Higher values make the $FA$ function consider very few, highly activated features, while lower values expand the search space. As the activations we considered as expressed as probability distributions, we use focus thresholding values in the range $[0, 1)$. All experiments are run with a fixed value of $\varepsilon=0.02$ and PGD attacks run for 5 iterations.

\begin{figure}
    \begin{subfigure}[t]{0.31\textwidth}
        \includegraphics[width=\columnwidth]{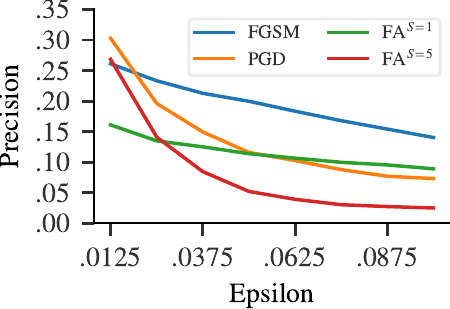}
        \caption{Effect of the $\varepsilon$-ball radius. Values are pixel colors in a [0,1] range.}
        \label{fig:explore_epsilon}
    \end{subfigure}
    \hfill
    \begin{subfigure}[t]{0.31\textwidth}
        \includegraphics[width=\columnwidth]{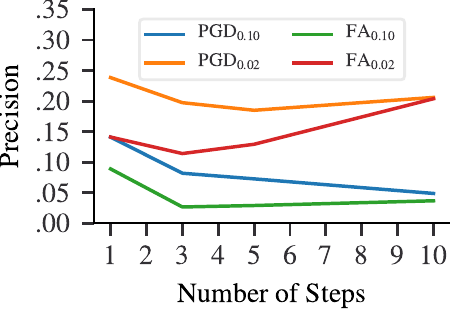}
        \caption{Effect of the perturbation granularity. The footers of the methods indicate the $\varepsilon$ radius.}
        \label{fig:explore_steps}
    \end{subfigure}
    \hfill
    \begin{subfigure}[t]{0.31\textwidth}
        \includegraphics[width=\columnwidth]{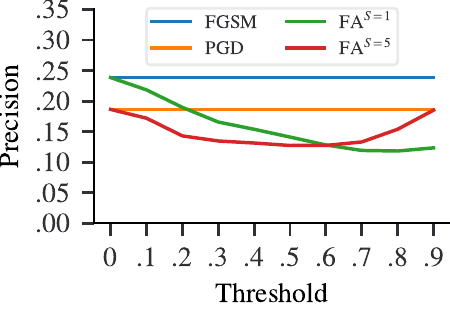}
        \caption{Effect of the focusing threshold. FGSM and PGD are not affected by this hyperparameter.}
        \label{fig:explore_focus}
    \end{subfigure}

    \caption{Effect of various hyperparameters on the precision of detections. The measure is mAP, and the baseline for Faster R-CNNis 0.469. \textit{FA} methods' footer is the number of iterations; the exponent is the focusing threshold.}
    \label{fig:exploratory_plots}
\end{figure}

Figure \ref{fig:explore_focus} shows the decrease in precision associated with different focus thresholds. The effectiveness of standard FGSM and PGD attacks is constant because the threshold does not affect them. PGD shows an increase in efficacy for focus values around 0.5, and worse performance as the threshold is raised or lowered. On the other hand, FGSM performs best with high focus threshold values (0.8 to 0.9), and actually exceeds PGD at a fraction of the computational cost when the threshold is above 0.6.

\subsection{Performance Analysis}
The execution time of our method compared to standard FGSM and PGD is presented in Table \ref{tab:speed} for different focus thresholding values. We run experiments on a single RTX 3070 GPU, and consider only computational time, therefore excluding I/O and main memory to GPU data movement. Results show that our focused adversarial attacks are either equivalent or marginally faster than the other methods. PGD performance is computed for 5 iterations of the algorithm.

\begin{table}
\caption{Performance of adversarial attack methods. FGSM and PGD show a single value because they are unaffected by the focus threshold. The superscripts for the FA algorithm indicate the number of iterations.}
\label{tab:speed}
\centering
\begin{tabular}{@{}lccccc@{}}
\toprule
            & \multicolumn{5}{c}{Focus Threshold}        \\
Attack      & 0.1    & 0.3    & 0.5    & 0.7    & 0.9    \\ \midrule
FGSM        & -      & -      & 139 ms & -      & -      \\
PGD         & -      & -      & 705 ms & -      & -      \\
FA$^1$      & 139 ms & 138 ms & 137 ms & 137 ms & \textbf{136 ms} \\
FA$^5$      & 691 ms & 683 ms & 677 ms & 675 ms & \textbf{672 ms} \\ \bottomrule
\end{tabular}
\end{table}

\section{Evaluation}
\label{sec:evaluation}

We evaluate the performance gain of focused adversarial attacks over FGSM and PGD in terms of efficacy and processing speed, with a brief insight on final perturbation magnitudes. Evaluations are carried out on two publicly available datasets: the full COCO 2017 validation split \cite{coco}, and Pascal VOC 2007 test \cite{pascal}.  These are publicly available datasets and have no personally identifiable information or offensive images.

\subsection{Models}
\label{sec:models}
Focused adversarial attacks can be carried out on any learned machine learning model. Though we suggest their use on models that output probabilities or probability distributions, with some tuning any model can compatible. The experiments included in this section are run on three state of the art object detection models: SSD300 \cite{ssd300}, Faster R-CNN \cite{frcnn}, and RetinaNet \cite{retinanet}. All models are trained on the COCO train 2017 \cite{coco} dataset.

\noindent 
\textbf{SSD300.}
The model and parameters of SSD300 are taken from Viet Nguyen's implementation\footnote{https://github.com/uvipen/SSD-pytorch/}. The model's backbone is a ResNet50\cite{resnet50} network. As the neural network's outputs are not normalized by either sigmoid or softmax functions, we manually apply a softmax layer to turn the class activations at each location into probability distributions.

\noindent
\textbf{RetinaNet.}
We use Yann Henon's implementation of RetinaNet\footnote{https://github.com/yhenon/pytorch-retinanet}. This model's backbone is also a ResNet50 architecture, and since its outputs are already probability distributions, no further tuning is required.

\noindent 
\textbf{Faster R-CNN.}
The implementation of Faster R-CNN is taken from Pytorch's torchvision library\footnote{https://pytorch.org/vision/stable/models.html}. The backbone is a Feature Pyramid Network-based (FPN) ResNet50, and we adjust its parameters when executing attacks to not filter out any detections and bounding boxes, so that the full network's output is used.

\subsection{Effectiveness}

Table \ref{tab:coco} shows the mAP scores of the three models on the COCO detection task for $\varepsilon$-balls of radius $0.1$ and $0.02$. In order to avoid overfitting the threshold to each particular model to artificially enhance results, we use a constant focusing threshold $t = 0.5$. In the first column we reproduce the officially reported precision scores for each model, and then report the mAP scores for various attacks. While both FGSM and PGD show a significant reduction in mAP, our attacks perform drastically better, even at minimal $\varepsilon$ values.

\begin{table}[htp]
\caption{mean Average Precision (mAP) on the COCO 2017 val dataset for different $\varepsilon$-ball radiuses.}
\label{tab:coco}
\centering
\begin{tabular}{@{}ll|cc|cc|cc|cc@{}}
    \toprule
    & & \multicolumn{4}{c|}{$\varepsilon = 0.1$} & \multicolumn{4}{c}{$\varepsilon = 0.02$} \\
    \multicolumn{2}{l|}{Model Baseline} & FGSM  & PGD   & FA$^1$ & FA$^5$ & FGSM  & PGD   & FA$^1$ & FA$^5$ \\ 
    \midrule
    RetinaNet & 0.345   & 0.198 & 0.097 & \textbf{0.065} & 0.074    & 0.288 & 0.216 & 0.081 & \textbf{0.020} \\
    SSD300    & 0.244   & 0.113 & 0.102 & 0.056 & \textbf{0.043}    & 0.188 & 0.168 & 0.088 & \textbf{0.076} \\
    F. R-CNN  & 0.469   & 0.162 & 0.067 & 0.078 & \textbf{0.028}    & 0.262 & 0.161 & \textbf{0.122} & 0.181 \\ 
\bottomrule
\end{tabular}
\end{table}

Table \ref{tab:pascal} compares our focused attacks with other current state-of-the-art adversarial attacks to fool object detectors. To fool SSD300 we use a threshold $T=0.1$, whereas for Faster R-CNN we set it to $T=0.5$; these values were obtained with a 10-step hyperparameter search and are supported by the fact that SSD300 is more sparsely activated than Faster R-CNN, and thresholding activations at $0.5$  often result in all activations being filtered out and the gradients reducing to $0$. Focused Attacks perform as well or better than other gradient-based attacks, and also provide equal or greater performance to GAN-based methods without requiring training any separate network.

\begin{table}[htp]
\caption{mAP resulting of adversarial attacks effectiveness on the Pascal VOC 2007 test set. PSNR metric is included to measure visual perturbation; it is in range $[0,100]$ and higher values are better. $T$ is the thre}
\label{tab:pascal}
\centering
\begin{tabular}{@{}ll|ccccc|cc@{}}
    \toprule
    & & \multicolumn{5}{c|}{Gradient Attacks} & \multicolumn{2}{c}{GAN Attacks} \\
    \multicolumn{2}{l|}{Model Baseline} 
            & FGSM  & PGD   & FA$^1$ & FA$^5$
            & DAG\cite{dag} & UEA\cite{uea} & MI \cite{mi} \\ 
    \midrule
    SSD300    & 0.686    & 0.564 & 0.531 & 0.200 & 0.048 & 0.640      & 0.200 & 0.160 \\
    F. R-CNN  & 0.778    & 0.431 & 0.388 & 0.204 & 0.056 & 0.050      & 0.050 & 0.060 \\ 
    \midrule
    \multicolumn{2}{c|}{PSNR} & 21 & 28 & 22 & 30 & 31  & 28 & 30 \\
    \bottomrule
\end{tabular}
\end{table}

\subsection{Speed}
\label{sec:speed}
The average adversarial attacks execution times are reported in Table \ref{tab:times}, and result from using a single RTX 3070 GPU. These times are not limited to the computational time, but also include I/O and main memory to GPU (and back) overhead. As this overhead is equivalently present in each of the three models, the results are comparable. In all cases, focused adversarial attacks are equally fast or faster than their baseline counterparts.

\subsection{Perceptibility}
Focused adversarial attacks are constrained by a pixel-wise upper bound $\varepsilon$. Within the related $\varepsilon$-ball, however, images may be more or less perturbed across all their pixels. As the $L_\infty$ distance metric is unable to capture this difference, we also show the $L_1$ perturbation magnitude across samples. Given two pixels, their $L_1$ distance tells how many values they are apart from each other, or their absolute difference.

For each image and its adversarial counterpart, $\mathbf{x}$, $\mathbf{x'} \in [0,1]^{W \times H \times 3}$, their mean $L_1$ distance is $\frac{\|\mathbf{x} - \mathbf{x'}\|_1}{W H C}$. In contrast with some works in the literature \cite{daedalus}, we use the \textit{mean} of the norm because of two reasons: it is more interpretable, and it is comparable across images of different sizes.

\begin{figure}
\centering
    \begin{minipage}{0.4\textwidth}
        \centering
        \includegraphics[width=\columnwidth]{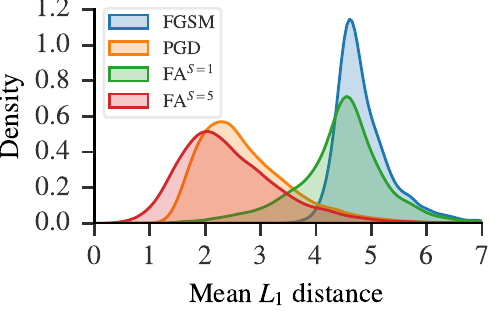}
        \caption{Distribution of $L_1$ norms of the perturbation masks generated with Faster R-CNN. The dashed lines indicate the mean of each distribution. Values on the abscissa are average pixel difference in the image, and refer to a $[0,255]$ range.}
        \label{fig:histograms}
    \end{minipage}
    \hspace{0.5cm}
    \begin{minipage}{0.55\textwidth}
        \centering
        \captionsetup{type=table}
        \caption{Average time to cloak a COCO image.}
        \label{tab:times}
        \begin{tabular}[b]{@{}lcc|cc@{}}
            \toprule
            Model     & FGSM  & PGD   & FA$^1$ & FA$^5$ \\ 
            \midrule
            RetinaNet & 130 ms & 645 ms & 117 ms  & 585 ms \\
            SSD300    & 95 ms  & 168 ms & 34 ms   & 167 ms \\
            F. R-CNN  & 183 ms & 919 ms & 171 ms  & 817 ms \\ 
        \bottomrule
        \end{tabular}
    \end{minipage}
\end{figure}

\begin{figure*}[t]
    \includegraphics[width=\textwidth]{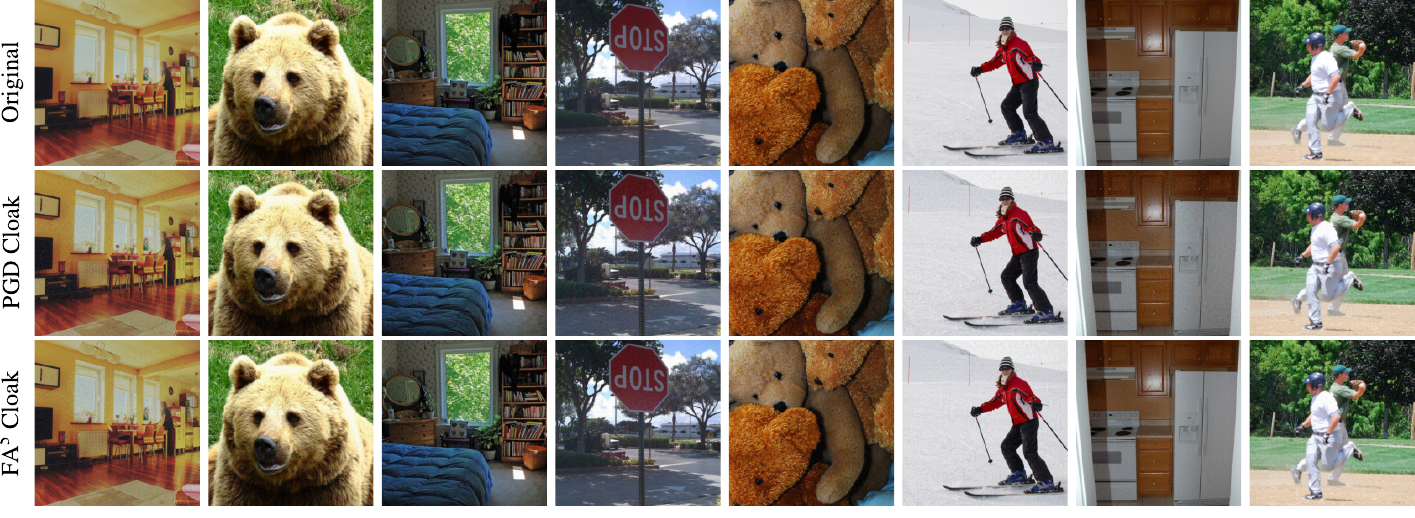}
    \caption{Adversarial examples generated with standard PGD and with our method. Cloaks are computed with Faster R-CNN, with 5 steps of gradient descent, and within an $\varepsilon$-ball of radius $\varepsilon=0.1$ on a $[0,1]$ scale.}
    \label{fig:samples}
\end{figure*}

Figure \ref{fig:histograms} shows the distribution of mean $L_1$ norms of the perturbation vectors generated with Faster R-CNN on the COCO 2017 validation dataset, using FGSM or 5 iterations of PGD. Our focused adversarial examples show a decrease of 10\% in the intensity of perturbations applied to images, under the same hyperparameter settings. We believe this is due to our perturbations being geared towards cloaking sensitive parts of images, while producing randomly-oriented changes in nearby pixels, thus leading to subsequent cloaks often canceling the previous ones out. Visual proof of this behavior is found in the examples in Figure \ref{fig:samples}.

Additionally, Table \ref{tab:pascal} includes a comparison of the Peak Signal to Noise Ratio (PSNR) of cloaked images across methods. The higher this value, the more clear is an image. Focused attacks perform better than baseline FGSM and PGD attacks, and equally well to GAN-based adversarial methods.

\section{Limitations and Future Works}
\label{sec:conclusion}

In this paper we propose \textit{focused adversarial attacks}, a gradient-based adversarial machine learning attack to break object detectors. By targeting adversarial perturbations only towards sensitive regions of images, focused attacks are more effective and also less visible than other state-of-the-art methods, under the same constraints. We believe this algorithm will be another important tool in ML practitioners' arsenal to evaluate the vulnerability of ML models and design more robust systems.

Being gradient-based, focused attacks require white-box access to the models they attack. While transferability has been shown in a number of similar works, it is usually constrained to special cases or requires significantly invasive perturbations to be effective. At the same time, ensemble attacks have shown good transferability and our system can easily be incorporated into an ensemble model.

In the future, we intend to optimize the implementation of focused attacks to reduce the processing time per frame down to at most $40ms$. This will allow us to perform real-time object cloaking in video feeds (at 24 FPS), and ideally adapt the attacks to the physical world.

\bibliographystyle{plain}
\bibliography{sources}

\end{document}